\documentclass[pdflatex,sn-mathphys-num]{sn-jnl}

\usepackage{graphicx}%
\usepackage{multirow}%
\usepackage{amsmath,amssymb,amsfonts}%
\usepackage{amsthm}%
\usepackage{mathrsfs}%
\usepackage[title]{appendix}%
\usepackage{xcolor}%
\usepackage{textcomp}%
\usepackage{manyfoot}%
\usepackage{booktabs}%
\usepackage{algorithm}%
\usepackage{algorithmicx}%
\usepackage{algpseudocode}%
\usepackage{listings}%

\theoremstyle{thmstyleone}%
\theoremstyle{thmstyletwo}%

\theoremstyle{thmstylethree}%

\begin{document}

\title[Article Title]{SolarFCD: A Large-Scale Dataset and Benchmark for Solar Fault Classification in Photovoltaic Systems}


\author[1]{\fnm{Misbah} 
\sur{Ijaz}}\email{25016119-001@uog.edu.pk}

\author*[2]{\fnm{Saif Ur Rehman} 
\sur{Khan}}\email{xuz72wot@rptu.de}
\author[1]{\fnm{Abd Ur} \sur{Rehman}}\email{a.rehman@uog.edu.pk}

\author[2]{\fnm{Arooj} \sur{Zaib}}\email{arooj.zaib@dfki.de}

\author[3,4]{\fnm{Sebastian} \sur{ Vollmer}}\email{sebastian.vollmer@dfki.de}
\author[2,3,4]{\fnm{Andreas} \sur{Dengel}}\email{andreas.dengel@dfki.de}

\author*[3,4,5]{\fnm{Muhammad Nabeel} \sur{Asim}}\email{muhammad\_nabeel.asim@dfki.de}

\affil[1]{\orgdiv{Department of Computer Science}, \orgname{University of Gujrat}, {\city{Gujrat}, \postcode{51700}, \country{Pakistan}}}

\affil[2]{\orgdiv{Department of Computer Science}, \orgname{Rhineland-Palatinate Technical University of
Kaiserslautern-Landau}, \orgaddress{\city{Kaiserslautern}, \postcode{67663}, \country{Germany}}}

\affil[3]{\orgname{German Research Center for Artificial Intelligence}, \orgaddress{\city{Kaiserslautern}, \postcode{67663}, \country{Germany}}}

\affil[4]{\orgname{Intelligentx GmbH (intelligentx.com)}, \orgaddress{\city{Kaiserslautern}, \country{Germany}}}


\affil[5]{\orgname{Department of Core Informatics, Graduate School of Informatics ,Osaka Metropolitan University}, \orgaddress{\city{Saka, 599-8531}, \country{Japan}}}


\abstract{The increasing global deployment of solar photovoltaic (PV) systems needs robust, scalable, and automated inspection technologies capable of detecting a wide range of panel flaws under a variety of operating situations. The lack of large-scale, multi-modal, publicly available annotated datasets is a major obstacle preventing advancement in this field. We introduce SolarFCD, an extensive dataset of solar panel defects created by methodically combining and reconciling three publicly accessible datasets covering two imaging modalities: RGB/Drone images and Thermal Infrared. The dataset consist of 4,435 images arranged under four unified defect classes such as: healthy images, Surface Obstruction, structural fault, and electrical fault. The dataset was divided into training, validation, and test splits at an 80:10:10 ratio through methodical label mapping, near-duplicate removal, and targeted augmentation of minority classes. Sixteen classification architectures from five design families were trained and assessed on the dataset to provide repeatable benchmark baselines. With an accuracy of 86.68\%, precision of 88.65\%, recall of 88.62\%, and F1-score of 88.17\%, ResNet101V2 performed the best overall. Per-class results showed balanced detection across all four defect categories within a narrow performance band of less than 1.2 percentage points. To promote open and repeatable research in automated PV inspection and solar energy operations and maintenance, the dataset, annotation files, and baseline code are made openly available.}





\maketitle

\section{Background \& Summary}\label{sec1}
The world's energy system is changing quickly as countries move away from fossil fuels. This shift is mostly being driven by renewable energy, especially solar and wind power, which is also a major factor in reducing climate change \cite{johnson2021stabilisation}. Solar power is a promising answer to the world's expanding electricity demand because of its low cost and versatility, which allow it to be used for both on and off grid applications in practically any place \cite{zhang2024balancing}. According to recent estimates from the International Energy Agency (IEA) \cite{iea_renewables_2025}, solar PV capacity additions are expected to reach 4600 gigawatts (GW) annually by 2030 in an expedited scenario. Over the next five years, solar PV capacity is expected to more than double, driving the rise of renewable energy globally.

The requirement for reliable and effective operation is expanding due to the increasing number of solar systems worldwide \cite{bamisile2025environmental}. 
A variety of environmental conditions can impact the longevity and performance of individual solar panels. Over time, panels may deteriorate due to dust buildup, physical damage, and weather related stresses such as hail, snow, and extremely high or low temperatures \cite{said2018effect}. These deteriorating elements can be especially severe in tough or severely contaminated locations, necessitating more frequent maintenance and repairs. The need for automated systems that can precisely evaluate the state of solar panels is highlighted by the labor intensive, error prone, and expensive nature of traditional manual inspections \cite{malek2023design}.

In order to improve energy production, lower maintenance costs, and prolong the lifespan of solar panels, automated systems that monitor solar panels in real time can optimize maintenance schedule and minimize personnel input \cite{das2018metaheuristic}. The creation of precise classification and diagnostic models is essential to optimizing maintenance since it reduces downtime, increases long-term system efficiency, and finds faults and performance problems early.

The difficulty of identifying and classifying different forms of degradation, such as dust buildup, microcracks, and physical damage, makes automated solar panel classification difficult \cite{bamisile2025environmental}. Conventional approaches sometimes fall short of providing scalable, real time solutions, especially when monitoring large-scale solar farms \cite{dhanraj2021effective}. They also have trouble spotting small problems that can eventually lower panel performance. Machine learning (ML) algorithms have been used to automate solar panel classification in response to these constraints \cite{wang2024locating}. Early methods depended on manually created feature extraction from images and on more basic ML techniques like support vector machines (SVM) and decision trees \cite{dhanraj2021effective}. Although these models made automation possible, they had poor accuracy and restricted scalability, especially when it came to handling the variations in solar panel conditions in various settings \cite{shaban2024detection, hossain2025enhancing}. Additionally, these models' capacity to adjust to changing deterioration patterns was constrained by their reliance on manually designed characteristics.

Deep learning (DL) developments, particularly convolutional neural networks (CNNs), have the potential to increase classification accuracy by directly learning features from unprocessed visual input \cite{jonathan2025multimodal, mehta2024utilizing}. But CNNs also struggle with generalization \cite{ejiyi2024maccom, tella2025solar}, especially when dealing with heterogeneous datasets that differ in terms of weather, lighting, and damage types. In order to better handle the intricacies of solar panel condition categorization, more advanced DL models \cite{araji2024utilizing} have been developed as a result of these difficulties. Despite these developments, many solar energy providers are still unable to use DL models because they still demand significant computational resources and big, well labeled datasets \cite{guo2023accurate}. 

The public resources that are currently available are each limited in one or more crucial aspects: they are limited to a single imaging modality, cover too few defect categories, offer coarse image-level labels that are inappropriate for object detection tasks, have too few annotated samples for trustworthy model generalization, or use annotation conventions that are inconsistent with those of other datasets \cite{chen2025open}. Due to this fragmentation, researchers are forced to either invest significant effort in preprocessing and harmonizing multiple incompatible sources independently with no guarantee of consistency, or train on limited, single-source data that does not reflect the full defect spectrum encountered in real inspections. 

To overcome these limitations, we present the Solar Fault Classification Dataset (SolarFCD), a large-scale, multimodal dataset of solar panel defects created by systematically combining, re-annotating, and harmonizing 3 publicly accessible datasets across two imaging modalities. For researchers and practitioners working on automated PV inspection, the dataset offers a single resource that supports object detection and image classification tasks across a comprehensive taxonomy of four defect categories.

\subsection*{Problem statement}
The public solar PV datasets that are currently available are dispersed and only cover a limited number of defect categories or imaging modalities. Surface obstruction, structural fault, and electrical faults across visual and thermal modalities are not all included in a single collection with uniform annotations. Researchers are forced to manually combine conflicting datasets as a result, which restricts model generalization. In order to close this gap, our study integrates three distinct public sources to provide a unified four-class dataset that allows for harmonized multi-class defect categorization for thorough PV inspection.

\subsection*{Research objectives}
The main goal of this work is to create and make available a unified, large-scale, multi-modal dataset in order to solve the data fragmentation issue in automated solar panel defect classification. The work is guided by the following precise objectives:
\begin{itemize}
    \item Combining several publicly available solar defect datasets from thermal and RGB modalities into a single, harmonized dataset.
    \item Creating a precise defect taxonomy with clear semantic concepts that covers all significant PV defect types.
    \item Utilizing expert validation and inter-annotator agreement assessment to standardize annotation format and quality.
    \item To create repeatable baselines, benchmark representative classification models.
    \item To encourage community-driven research and extension, all data, pipeline code, and documentation will be made available under an open license.
\end{itemize}

\subsection*{Contribution}
\begin{itemize}
    \item \textbf{A large-scale multi-modal dataset:} We provide SolarFCD, a comprehensive multi-modal collection of 4,435 annotated images from three public sources that span thermal and RGB modalities.
    \item \textbf{A comprehensive and standardized defect taxonomy:} We establish detailed -class defect taxonomy that includes expert validated label mappings across all source datasets and exact semantic descriptions.
    \item \textbf{High-quality image-level annotations:} High-quality image-level annotations are supplied for every image, facilitating tasks related to classification.
    \item \textbf{Reproducible baseline benchmarks:} Sixteen classification architectures from five design families, DenseNet, EfficientNetV2, MobileNetV2, ResNet, VGG, and Vision Transformer, were trained and assessed on the dataset to provide repeatable benchmark baselines. 
    \item \textbf{A fully documented and reproducible construction pipeline:} The entire building pipeline, including taxonomy mappings, annotation protocols, and preprocessing scripts, is made available for community development and independent verification.
    \item \textbf{Open distribution under a permissive license:} Lastly, the complete dataset and documentation are freely available under [CC BY 4.0].
\end{itemize}

\section{Critical analysis of existing public datasets}
Over the past ten years, there has been an increasing body of work on the automated detection of flaws in PV modules, and several datasets have been made publicly available to support this effort. However, an in-depth look at the datasets included in SolarFCD shows that, despite each resource's value within its specified scope, each has one or more structural flaws that make it unsuitable as a stand-alone training source for reliable, deployment ready inspection models. This section's analysis assesses each dataset in four crucial areas: dataset name, dataset scale, defect taxonomy breadth, and imaging modality coverage. Structured summaries of these features are given in \ref{tab:existing_datasets}. 
\begin{table}[h!]
\caption{Overview of existing public datasets}\label{tab:existing_datasets}%
\begin{tabular}{@{}llllll@{}}
\toprule
Author & Dataset Name & Modality & Images & Defect classes\\
\midrule
\cite{dust-dataset} & Solar Panel Dust Detection & RGB only & 2,562 images & 2 classes\\
\cite{RGBdataset} & Solar Panel Clean \& Faulty Images & RGB only & 873 images & 6 classes \\
\cite{Pvmd} & PVMD dataset & Thermal Infrared & 1,000 images & 3 classe \\
\botrule
\end{tabular}
\end{table}

With 2,562 images, the Dust Detection Dataset \cite{dust-dataset} is the biggest single contributor to SolarFCD. Its clean and dusty subcategories offer robust coverage of the Healthy and Surface Obstruction classes. The dataset's primary restriction is its deliberate focus; it was created solely to address soiling detection and does not include any examples of electrical or structural damage. A model trained solely on this source would achieve excellent accuracy in clean versus dusty discrimination but would be completely ignoratnt to the two most operationally significant defect categories: structural and electrical problems, which offer the highest risk of module failure and fire hazard. Because it covers two of the four target classes and one visible light modality, its contribution to the combined dataset is valuable but strictly limited.

The solar panel clean and faulty dataset \cite{RGBdataset} is the most taxonomically varied of the three sources, covering six original subcategories that map across all four unified classes. But for trustworthy DL training, its per-class sample counts are woefully inadequate: Only 84 and 97 images, respectively, are contributed by physical and electrical damage; these counts are much below the minimal requirement for learning stable discriminative features without significant overfitting. Because no DL architecture can effectively generalize from fewer than 100 examples per class without significant augmentation, the dataset's coverage of structural and electrical defect classes is present in taxonomy but absent in practical utility. In addition, as all of the data are visible light RGB, there is no thermal modality coverage for the hotspot and resistive fault signals, which can only be detected thermally.

The PVMD dataset \cite{Pvmd} fills the modality gap left completely open by both other sources and provides defect instances in SolarFCD by contributing 1,000 images solely through infrared acquisition. It is the main source of thermal ground truth for the two most safety-critical defect categories because its Cracks and Hotspots subcategories translate directly to the structural fault and electrical fault classes, respectively. But it has serious limits. The dataset cannot support classifier training in isolation because it only covers three defect classes and does not include examples of healthy or surface obstruction. The thermal subset's near complete collapse following duplicate removal reveals a basic quality flaw in the original dataset that would have gone unnoticed if strict deduplication had not been part of the compilation process.

No single dataset simultaneously achieves a large image count, a comprehensive defect taxonomy, and broad modality coverage, as shown in Table \ref{tab:existing_datasets} across all columns. Resources with a large number of images are typically single modality and taxonomically limited, whereas resources with a wider coverage of defects are typically smaller. This inverse link between breadth and scale, which results from the practical challenge and expense of obtaining and annotating multi-modal data at scale, is a structural characteristic of the current environment rather than an accidental gap. This is addressed by SolarFCD, which aggregates data from multiple sources rather than relying on a single data gathering effort.

\section{Dataset construction pipeline}
\begin{figure}[h!]
    \centering
    \includegraphics[width=0.9\linewidth]{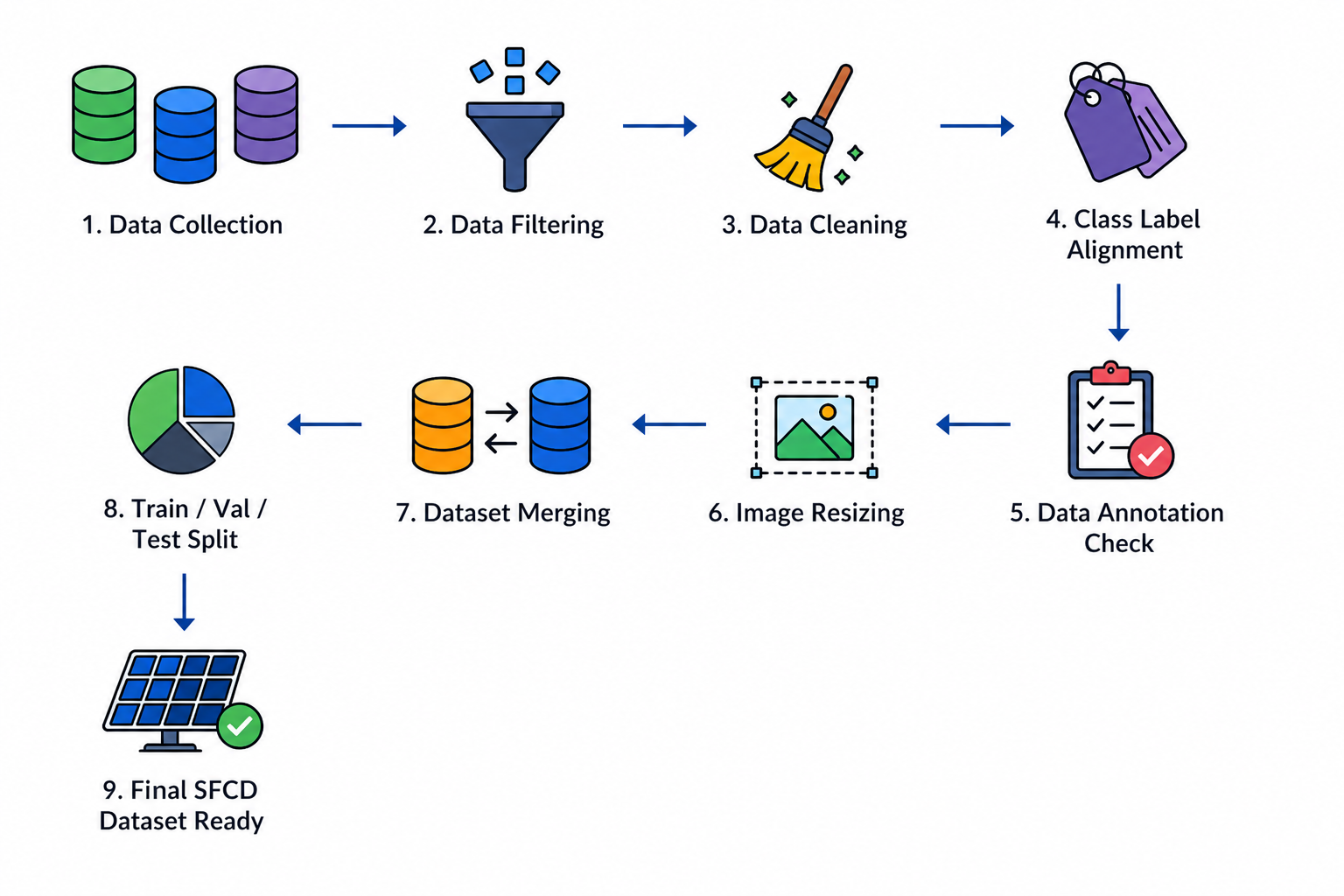}
    \caption{Workflow for SolarFCD Dataset}
    \label{fig:workflow}
\end{figure}

A methodical, multi-stage pipeline was used to build SolarFCD in order to guarantee consistency, repeatability, and traceability throughout all processes, from source acquisition to final release. The pipeline's architectural overview is shown in Figure \ref{fig:workflow}. We built the SolarFCD dataset by combining 3 publicly accessible datasets. Each source dataset, its original annotations, and the preparation done before merging are described in section \ref{sec4}.
\subsection{Dataset acquisition and selection criteria}
Building a unified multimodal SolarFCD dataset from publicly accessible sources requires a methodical acquisition approach that goes beyond simply downloading all available resources. The three datasets included in SolarFCD were selected through a methodical procedure that evaluated each dataset based on factors such as dataset size, image acquisition technique, and the contribution of the imaging modality. This section explains the particular features of the obtained sources in each of these three dimensions and records the reasoning behind each selection choice.

To train DL models that consistently generalize across a variety of realworld inspection settings, scale is a fundamental prerequisite. Models that either overfit to the training distribution or are unable to learn discriminative features for minority defect classes are produced by datasets that are too small to give sufficient per-class sample counts. On the other hand, class imbalance is introduced by a dataset that is large overall but concentrated in a single defect category, which biases model predictions toward the majority class at the expense of less common but operationally crucial fault kinds.

The authenticity and domain variety of the resulting training data are directly impacted by the imaging conditions under which a dataset was obtained. Three general methods are used to obtain solar defect datasets in the literature: controlled laboratory acquisition, structured field acquisition, and drone-based survey, or images were scraped from the internet. A combined dataset that incorporates all three approaches benefits from a wider diversity of imaging situations than any one approach can offer, and each methodology generates data with unique statistical characteristics. Table \ref{tab:acquisition-technique} summarizes the collection startegy and common equipment associated with each source.
\begin{table}[h!]
\caption{summary of the acquisition techniques}
\label{tab:acquisition-technique}
\begin{tabular}{lll}
\toprule
\textbf{Data source} & \textbf{Collection startegy}  & \textbf{Common Equipment}                   \\
\midrule
\cite{dust-dataset, RGBdataset} & Web scraping                  & Mixed (phones, web images, unknown cameras) \\
\cite{Pvmd} & experimental field collection & DJI Mavic 3 Thermal Drone                \\
\botrule
\end{tabular}
\end{table}

\begin{figure}[h!]
    \centering
    \includegraphics[width=0.5\linewidth]{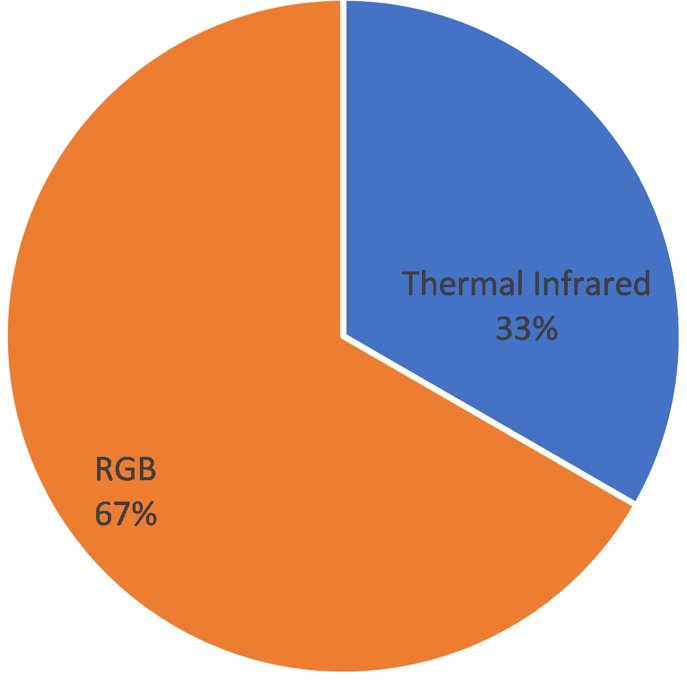}
    \caption{Imaging modality of source datasets}
    \label{fig:image-modality}
\end{figure}

Visible-light RGB imaging and thermal infrared imaging are the two main imaging modalities used in PV inspection that are covered by the source datasets included in SolarFCD. The visual appearance of typical images from each modality is shown in Figure \ref{fig:image-modality}, which highlights the fundamentally distinct feature spaces that RGB, thermal, and EL cameras expose for the same underlying fault situations. Since electroluminescence (EL) imaging is the main technique for identifying microcracks and cell-level electrical mismatch patterns that neither RGB nor thermal cameras can consistently resolve, the lack of EL imaging among the three source datasets is a noted limitation of the current version of SolarFCD.

\subsection{Annotation standardization}
The original annotation techniques used by the source datasets varied; normalization was necessary before integration. Each source dataset's annotation technique and the transformation employed during integration into SolarFCD are compiled in Table \ref{tab:annotation-methods}.
\begin{table}[h!]
\caption{Source annotation methods}
\label{tab:annotation-methods}
\begin{tabular}{lll}
\toprule
\textbf{Source Dataset} & \textbf{Original Annotation} & \textbf{Annotation Type} \\
\midrule
\cite{dust-dataset} & Binary class label per image & Image-level              \\
\cite{RGBdataset}      & Six-class label per image    & Image-level              \\
\cite{Pvmd}       & Three-class label per image    & Image-level             \\
\botrule
\end{tabular}
\end{table}

Instead of object-level bounding boxes or segmentation masks, all three source datasets provide image level classification labels, which are in line with the image classification job scope of SolarFCD. Since the dataset is intended to facilitate whole image defect classification rather than localized object recognition, no spatial re-annotation was necessary. A single integer class label per image that corresponds to the four-class taxonomy is the unified annotation format used for all images. These labels are kept in a CSV metadata file with columns for the image path, unified class label, source dataset identification, original sub-category name, and split assignment.

\subsection{License and usage terms}
SolarFCD is released under the Creative Commons Attribution 4.0 International (CC BY 4.0) license, permitting unrestricted use, redistribution, and adaptation for both research and commercial purposes provided that appropriate credit is given to the original dataset paper and source datasets. The content may be shared, duplicated, and expanded upon by users in any format or medium without first obtaining the authors' consent.
The three source datasets contained into SolarFCD each have their own distinct licenses, and users of this dataset are subject to the restrictions of the most restrictive applicable source license for each subset of images obtained from that source. Each source dataset's license and any usage limitations that apply to the relevant subset inside SolarFCD are compiled in Table \ref{tab: source-dataset-license}.

\begin{table}[h!]
\caption{Source dataset licenses and usage restrictions}
\label{tab: source-dataset-license}
\begin{tabular}{lll}
\toprule
\textbf{Original License} & \textbf{Redistribution Permission}      & \textbf{Source Dataset} \\
\midrule
CC0: Public Domain        & Allowed without restrictions            & \cite{dust-dataset}                        \\
Unknown                   & Cannot be assumed; permission uncertain &  \cite{RGBdataset}                       \\
CC BY 4.0                 & Allowed with proper attribution         &    \cite{Pvmd} \\
\botrule
\end{tabular}
\end{table}

\subsection{Data refinement and formatting}
Before being merged into a single training resource, raw pictures obtained from three source datasets needed to undergo a systematic refinement process. Four consecutive concerns were addressed by this process: eliminating redundant images, correcting class imbalance, standardizing annotation format, and creating consistent dataset splits. To guarantee complete reproducibility, each step is detailed below.

\section{Dataset statistics} \label{sec4}
Three publicly accessible source datasets, including two imaging modalities and several original defect subcategories, provided the raw image pool that was used to create SolarFCD. Each source dataset utilized its own internal labeling system before any preprocessing or harmonization, which did not exactly match the uniform four-class taxonomy employed in this work. As a result, the construction process was carried out in two steps: first, the raw images were categorized based on their source partitions; second, each original subcategory label was mapped, via expert review, to one of the four unified defect classes: healthy, surface obstruction, structural fault, and electrical fault.

\begin{figure} [h!]
    \centering
    \includegraphics[width=0.9\linewidth]{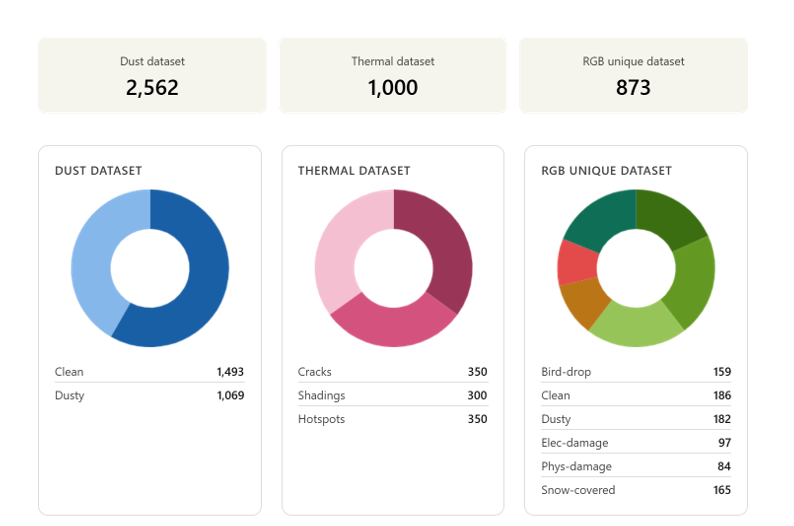}
    \caption{Raw source partition before mapping}
    \label{fig:Original_dataset}
\end{figure}

Before any label mapping, Figure \ref{fig:Original_dataset} shows the raw image counts for each source dataset. The 2,562 images from the dust dataset\cite{dust-dataset} were divided into two initial subcategories: dusty panels (1,069 images) and clean panels (1,493 images). Clean (186), Bird-drop (159), Dusty (182), Snow-Covered (165), Physical-Damage (84), and electrical damage (97) are the six original subcategories that the RGB dataset \cite{RGBdataset} contributed 873 images to, representing the wider variety of surface level defects recorded by visible-light camera acquisition. The 1,000 photos from the thermal dataset \cite{Pvmd} were divided into three thermally acquired subcategories: hotspots (350), cracks (350), and shadings (300), each of which was obtained using infrared photography while the module was operating actively. Before any label transformation was done, 4,435 images made up the aggregate raw pool from all three source datasets.

\subsection{Unified class mapping}
\begin{figure}[h!]
    \centering
    \includegraphics[width=0.9\linewidth]{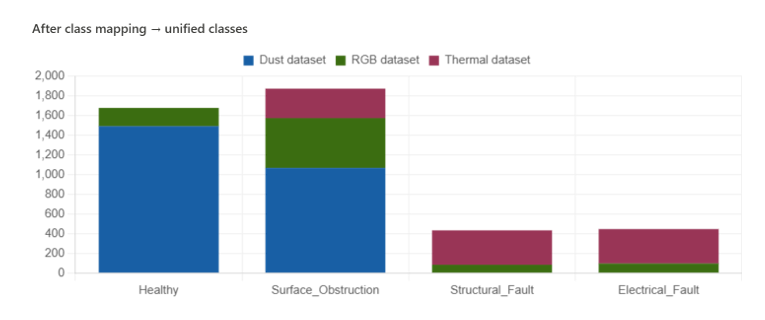}
    \caption{Class frequency visualization for the SolarFCD}
    \label{fig:class_mapping}
\end{figure}

Expert opinion was used to assign each original subcategory to one of the four unified taxonomy classes after the source partition evaluation. A total of 1,679 healthy panel images were produced by mapping clean images from the dust dataset \cite{dust-dataset} and RGB dataset \cite{RGBdataset} to the healthy class. The RGB Dataset's \cite{RGBdataset} bird-drop, dusty, and snow covered subcategories and the Thermal Dataset's \cite{Pvmd} shadings subcategory were combined into the Surface Obstruction class, which included 1,875 images and reflected the common physical mechanism of external material obstructing incident irradiance. For a total of 434 pictures, physical damage from the RGB dataset \cite{RGBdataset} and cracks from the thermal dataset \cite{Pvmd} were integrated under the structural fault class. A total of 447 images were mapped to the electrical fault class using electrical damage from the RGB dataset \cite{RGBdataset} and hotspots from the thermal dataset \cite{Pvmd}. 
Table \ref{tab:unified_class-mapping} summarizes the unified class mapping for the SolarFCD and gives the precise count and percentage for each category. This distribution is correspondingly shown in Figure \ref{fig:class_mapping}, which provides a visual comparison of the dataset's balance across several fault types.

\begin{table}[h!]
\caption{Unified class distribution after mapping}
\label{tab:unified_class-mapping}
\begin{tabular}{lllll}
\toprule
\textbf{Unified    Class} & \textbf{Dust    Dataset} & \textbf{RGB    Dataset}                          & \textbf{Thermal    Dataset} & \textbf{Raw    Total} \\
\midrule
Healthy                   & Clean (1493)             & Clean (186)                                      & —                           & 1679                  \\
Surface\_Obstruction      & Dusty (1069)             & Bird-drop (159), Dusty (182), Snow-Covered (165) & Shadings (300)              & 1875                  \\
Structural\_Fault         & —                        & Physical-Damage (84)                             & Cracks (350)                & 434                   \\
Electrical\_Fault         & —                        & Electrical-damage (97)                           & Hotspots (350)              & 447                   \\
\textbf{Total}            & \textbf{2562}            & \textbf{873}                                     & \textbf{1000}               & \textbf{4435}      \\  
\botrule
\end{tabular}
\end{table}

\begin{figure}[h!]
    \centering
    \includegraphics[width=0.9\linewidth]{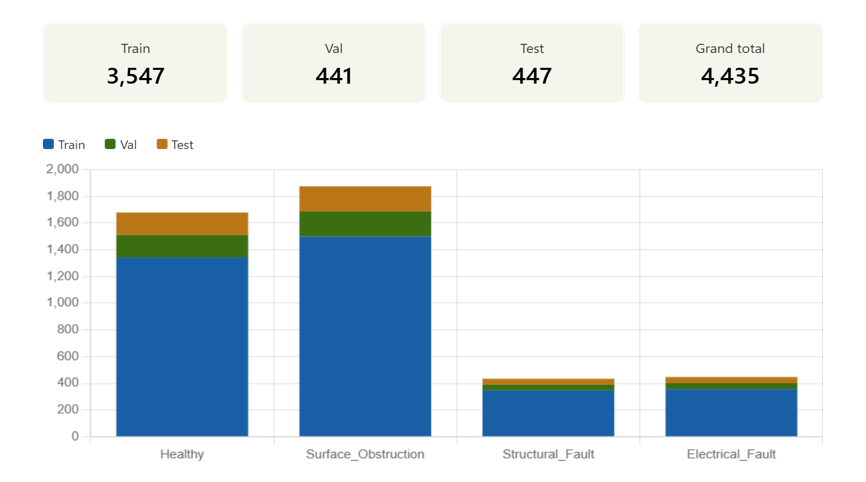}
    \caption{Unified class distribution split after mapping}
    \label{fig:class_mapping_split}
\end{figure}

After mapping, the dataset was divided into training, validation, and test partitions using an 80:10:10 ratio. The resulting class distribution is shown in Table \ref{tab:split_class_distribution}. There was a clear class disparity in the mapped distribution: With 1,875 images, surface obstruction accounted for the biggest share of the sample (42.3\%), followed by healthy (37.9\%) with 1,679 images. Structural faults and electrical faults, on the other hand, were significantly underrepresented (9.8\% and 10.1\%, respectively, with only 434 and 447 images). In comparison to the relatively uncommon structural and electrical fault instances available from public thermal and RGB inspection datasets, this imbalance is directly caused by the natural abundance of soiling and clean panel examples in field-collected visible-light sources. This imbalance required the targeted duplicate removal and augmentation interventions detailed in the following sections. Table \ref{tab:split_class_distribution} shows the precise sample counts for each set to illustrate the consistency of class representation following partitioning, and Figure \ref{fig:class_mapping_split} shows a comparison visualization of the data distribution.
\begin{table}[h!]
\caption{Class distribution after mapping with split counts}
\label{tab:split_class_distribution}
\begin{tabular}{llllll}
\toprule
\textbf{Class}       & \textbf{Train} & \textbf{Val} & \textbf{Test} & \textbf{Total} & \textbf{\% of dataset} \\
\midrule
Healthy              & 1,343          & 167          & 169           & 1,679          & 37.9\%                 \\
Surface\_Obstruction & 1,500          & 187          & 188           & 1,875          & 42.3\%                 \\
Structural\_Fault    & 347            & 43           & 44            & 434            & 9.8\%                  \\
Electrical\_Fault    & 357            & 44           & 46            & 447            & 10.1\%                 \\
\textbf{Total}       & \textbf{3,547} & \textbf{441} & \textbf{447}  & \textbf{4,435} & \textbf{100\%}  \\
\botrule
\end{tabular}
\end{table}

\subsection{Duplicate removal}
\begin{figure}[h!]
    \centering
    \includegraphics[width=0.9\linewidth]{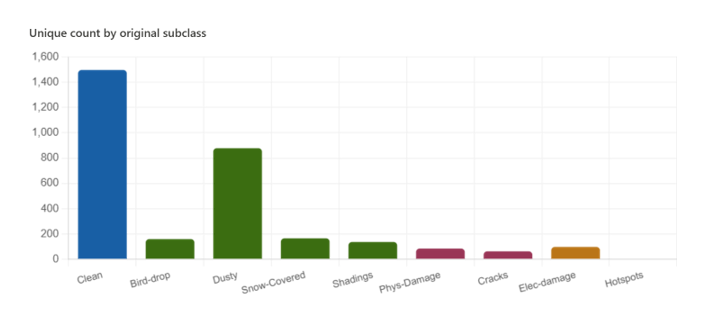}
    \caption{Unique Count by Original Subclass}
    \label{fig:unique_count_subclasses}
\end{figure}

\begin{figure}[h!]
    \centering
    \includegraphics[width=0.9\linewidth]{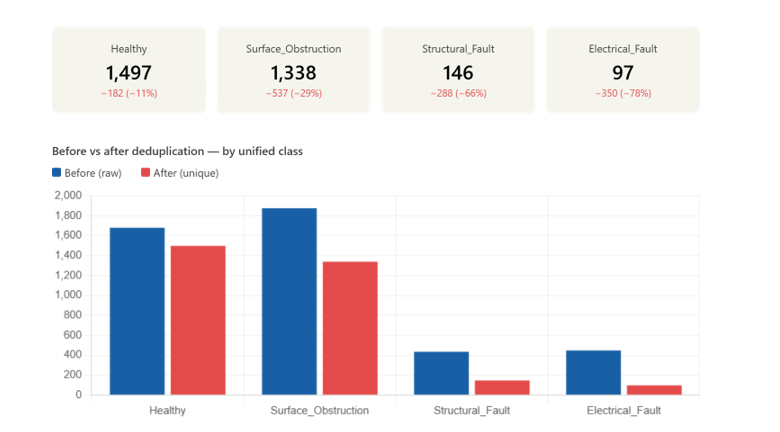}
    \caption{Comparative analysis of class distributions before and after the deduplication process}
    \label{fig:duplicate_removal}
\end{figure}

All subclasses were subjected to near-duplicate detection using perceptual hashing in order to find and eliminate redundant images that can falsely inflate model evaluation metrics by including nearly identical instances in both training and test splits. The per-subclass duplicate elimination result is shown in Table \ref{tab:duplicate_removal}. The Cracks subclass saw the biggest reductions, with only 62 unique instances remaining after 288 of 350 thermal images were determined to be near-duplicates and eliminated. This reduction of 82.3\% reflects the low visual diversity typical of thermally acquired crack imagery from a single installation source. The Shadings subclass lost 164 images, from 300 to 136, and the Dusty subclass lost 373 images, from 1,251 to 878 unique examples.
All 350 thermal hotspot images were found to be duplicates of entries existing in the pool, and the hotspots subclass was eliminated. There is more natural visual diversity among subclasses like bird drop, Snow Covered, physical damage, and electrical damage, which kept all images with no duplication found. 3,078 unique images made up the entire dataset after duplicates were eliminated, which is a 30.6\% decrease from the initial raw pool of 1,357 images. 
A rigorous deduplication pipeline was used to ensure the integrity of the model training process; Figures \ref{fig:unique_count_subclasses} and \ref{fig:duplicate_removal} show the breakdown of unique samples by their original subclasses and the shifts in class distribution, respectively, while Table \ref{tab:duplicate_removal} records the precise counts of removed instances.

\begin{table}[h!]
\caption{Summary of the duplicate removal process}
\label{tab:duplicate_removal}
\begin{tabular}{lllll}
\toprule
\textbf{Subclass} & \textbf{Unified class} & \textbf{Raw}   & \textbf{Unique} & \textbf{Removed} \\
\midrule
Clean             & Healthy                & 1,679          & 1,497           & $-$182             \\
Bird-drop         & Surface\_Obstruction   & 159            & 159             & 0                \\
Dusty             & Surface\_Obstruction   & 1,251          & 878             & $-$373             \\
Snow-Covered      & Surface\_Obstruction   & 165            & 165             & 0                \\
Shadings          & Surface\_Obstruction   & 300            & 136             & $-$164             \\
Physical-Damage   & Structural\_Fault      & 84             & 84              & 0                \\
Cracks            & Structural\_Fault      & 350            & 62              & $-$288             \\
Electrical-damage & Electrical\_Fault      & 97             & 97              & 0                \\
Hotspots          & Electrical\_Fault      & 350            & 0               & $-$350             \\
\multicolumn{2}{l}{\textbf{Total}}         & \textbf{4,435} & \textbf{3,078}  & \textbf{$-$1,357} \\
\botrule
\end{tabular}
\end{table}

\subsection{Augmentation strategy}
\begin{figure}[h!]
    \centering
    \includegraphics[width=0.9\linewidth]{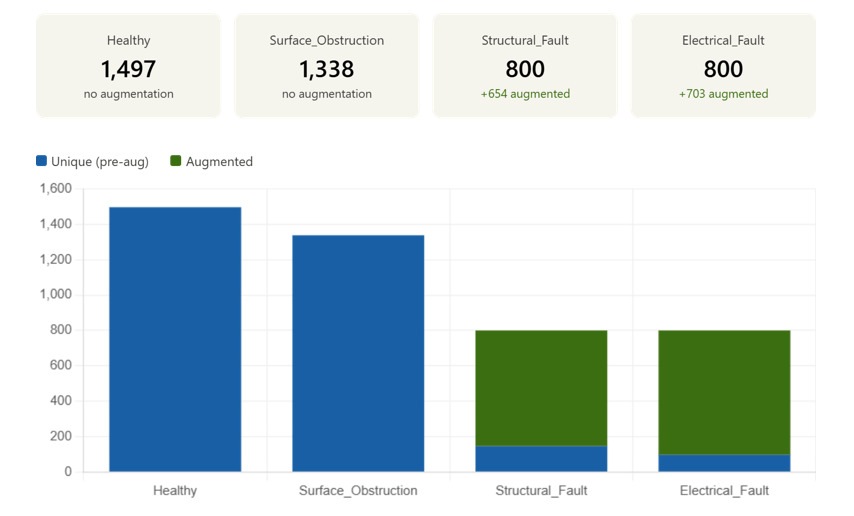}
    \caption{Comparative visualization of class distribution before and after targeted augmentation}
    \label{fig:augmentation}
\end{figure}

Due to the disproportionate reduction of the two minority classes, the duplicate elimination stage made the class disparity worse. Both classes were significantly underrepresented in comparison to the majority classes, with only 146 distinct images for structural fault and only 97 for electrical fault. Targeted augmentation was only administered to the two minority classes in order to restore equilibrium and provide enough training signal for dependable minority class learning. The augmentation pipeline used a combination of geometric transformations and photometric adjustments, such as horizontal and vertical flipping, random rotations between $-30^{\circ}$ and $+30^{\circ}$, brightness scaling (factor of 0.8 to 1.2), and Gaussian noise injection. These methods varied the minority samples' spatial and illumination features while maintaining methodological consistency.
By adding 654 more structural fault and 703 more electrical fault images, we increased the total number of images in each minority class to 800. The unique image counts of 1,497 and 1,338 for the Healthy and Surface Obstruction majority groups were unaltered. With 4,435 total photos and class proportions of 34.0\% healthy, 30.4\% surface obstruction, 18.2\% structural fault, and 18.2\% electrical fault, the final dataset following augmentation showed a significantly more balanced distribution than the pre-augmentation state. A thorough analysis of the minority class augmentation procedure, including the amount of synthetic samples produced by targeted augmentation, the starting unique counts, and the final balanced class totals shown in Table \ref{tab:augmentation}. The restoration of data balance for the structural and electrical fault categories is highlighted in Figure \ref{fig:augmentation}, which compares the distribution of classes before and after targeted augmentation.

\begin{table}[h!]
\caption{Detailed breakdown of augmentation process}
\label{tab:augmentation}
\begin{tabular}{lllll}
\toprule
\textbf{Class}       & \textbf{Unique} & \textbf{Augmented} & \textbf{Final total} & \textbf{\% of dataset} \\
\midrule
Healthy              & 1,497           & —                  & 1,497                & 34.0\%                 \\
Surface\_Obstruction & 1,338           & —                  & 1,338                & 30.4\%                 \\
Structural\_Fault    & 146             & +654               & 800                  & 18.2\%                 \\
Electrical\_Fault    & 97              & +703               & 800                  & 18.2\%                 \\
\textbf{Total}       & \textbf{3,078}  & \textbf{+1,357}    & \textbf{4,435}       & \textbf{100\%}        \\
\botrule
\end{tabular}
\end{table}

\subsection{Dataset distribution}
The expanded dataset was divided into 80:10:10 portions for training, validation, and testing. The resulting image counts for each class, and the splits are shown in Table \ref{tab:dataset_distribution}. 3,548 images make up the training split, of which 1,198 are healthy, 1,070 are surface obstructions, and 640 are each for structural and electrical faults. Each of the four classes is represented by 444 images in the validation and test splits, with 150 and 149 healthy images, 134 for surface obstruction, and 80 for structural fault and electrical fault. In addition to ensuring that training convergence monitoring through validation loss reflects performance across the entire defect taxonomy rather than being dominated by majority class dynamics, equal representation of both minority classes in all three splits guarantees that per-class evaluation metrics on the validation and test sets are computed over meaningful sample counts rather than trivially small populations.

\begin{table}[h!]
\caption{Final dataset distribution}
\label{tab:dataset_distribution}
\begin{tabular}{lllll}
\toprule
\textbf{Class}       & \textbf{Train} & \textbf{Validate} & \textbf{Test} & \textbf{Total} \\
\midrule
Healthy              & 1198           & 150               & 149           & 1497           \\
Surface\_Obstruction & 1070           & 134               & 134           & 1338           \\
Structural\_Fault    & 640            & 80                & 80            & 800            \\
Electrical\_Fault    & 640            & 80                & 80            & 800     \\
\botrule
\end{tabular}
\end{table}

\section{Technical validation}
The purpose of the technical validation phase is to evaluate the proposed dataset's robustness and generalizability. This section offers a quantitative assessment of the system's accuracy in classifying solar faults under defined standardized performance metrics by comparing the model to the partitioned training, validation, and test sets.
\subsection{Baseline results}
The classification performance of each of the sixteen architectures on the test split of SolarFCD is shown in Table \ref{tab:baseline_results}.
\begin{table}[h!]
\caption{Baseline classification results on SolarFCD}
\label{tab:baseline_results}
\begin{tabular}{lllll}
\toprule
\textbf{Methods}     & \textbf{Accuracy (\%)} & \textbf{Precision (\%)} & \textbf{Recall (\%)} & \textbf{F1-Score (\%)} \\
\midrule
DenseNet121          & 80.81                  & 0.8149                  & 0.8174               & 0.8140                 \\
DenseNet169          & 83.52                  & 0.8421                  & 0.8552               & 0.8455                 \\
DenseNet201          & 83.75                  & 0.8437                  & 0.8579               & 0.8490                 \\
EfficientNetV2B0     & 75.62                  & 0.7636                  & 0.7816               & 0.7709                 \\
EfficientNetV2B1     & 71.20                  & 0.7213                  & 0.7374               & 0.7240                 \\
EfficientNetV2B2     & 81.19                  & 0.8170                  & 0.8328               & 0.8231                 \\
MobileNetV2          & 79.01                  & 0.7963                  & 0.8003               & 0.7975                 \\
ResNet50V2           & 84.87                  & 0.8631                  & 0.8711               & 0.8660                 \\
\textbf{ResNet101V2} & \textbf{86.68}         & \textbf{0.8865}         & \textbf{0.8862}      & \textbf{0.8817}        \\
ResNet152V2          & 85.55                  & 0.8753                  & 0.8678               & 0.8711                 \\
VGG16                & 73.75                  & 0.7464                  & 0.7620               & 0.7524                 \\
VGG19                & 76.97                  & 0.7856                  & 0.7835               & 0.7844                 \\
VITB16               & 83.75                  & 0.8489                  & 0.8552               & 0.8484                 \\
VITB32               & 80.81                  & 0.8331                  & 0.8256               & 0.8292                 \\
VITL16               & 84.87                  & 0.8622                  & 0.8677               & 0.8598                 \\
VITL32               & 82.84                  & 0.8350                  & 0.8502               & 0.8393    \\
\bottomrule
\end{tabular}
\end{table}
With classification accuracy ranging from 15.48 percentage points from the lowest to the highest performer, Table \ref{tab:baseline_results}'s results show that SolarFCD facilitates significant and differentiated learning across all sixteen architectures. A useful benchmark requires that the dataset be sufficiently difficult to distinguish between model architectures of different capacities, which is confirmed by this dispersion.

Among all evaluated models, ResNet101V2 achieves the highest performance across all four metrics, with an accuracy of 86.68\%, precision of 88.65\%, recall of 88.62\%, and F1-score of 88.17\%. ResNet101V2 is the strongest baseline on SolarFCD according to this result, and its stated metrics are the main point of comparison going forward. The ResNet family with residual connections and V2 pretraining is very well matched to the feature distributions found in this dataset, as evidenced by the closely related ResNet152V2 achieving the second highest F1-score at 87.11\% and ResNet50V2 following with 86.60\%. The minor decrease in accuracy from ResNet101V2 to ResNet152V2 indicates that the dataset has sufficient complexity to benefit from increasing model depth up to the ResNet101V2 level. This is supported by the consistent performance ordering within the ResNet family, where deeper models typically outperform shallower ones.

DenseNet201 and DenseNet169 achieve F1-scores of 84.90\% and 84.55\%, respectively, which are marginally lower than the ResNet architectures but significantly higher than the EfficientNetV2 and VGG families, demonstrating competitive midrange performance. The feature reuse requirements of defect classification, where several layers of abstraction contribute complimentary information, seem to be well suited to DenseNet's dense connectivity structure. With an F1-score of 85.98\%, which puts ViT-L/16 among the best models, Vision Transformer topologies yield results that are generally equivalent to DenseNet. Given that transformers are typically thought to require bigger training datasets than are customary for domain specific defect benchmarks, the performance of ViT variations in comparison to convolutional architectures on this dataset is notable. Their competitive results indicate that SolarFCD's size and ImageNet pretraining are adequate for successful transformer fine-tuning in this field.

Among the assessed architecture families, the EfficientNetV2 family yields the most inconsistent results; EfficientNetV2B2 achieves a respectable F1-score of 82.31\%, while EfficientNetV2B1 records the lowest overall accuracy at 71.20\%. Since EfficientNetV2B2 has fewer parameters than a number of lower-performing models, this diversity within a single architecture family is an informative feature of the dataset rather than just a reflection of model capability. It implies that architecture selection for this dataset cannot be reduced to a straightforward capacity argument and that the compound scaling technique of EfficientNet interacts non-monotonically with the particular image statistics and class structure of SolarFCD. With VGG16 obtaining the lowest F1-score at 75.24\%, the VGG family regularly performs worse than contemporary designs. This is to be expected given its lack of residual connections and rather straightforward architectural design in comparison to the other families assessed.

MobileNetV2 outperforms VGG16, VGG19, EfficientNetV2B0, and EfficientNetV2B1 with a competitive accuracy of 79.01\% and F1-score of 79.75\%, despite its lightweight design meant for resource-constrained deployment. For edge deployment scenarios where a trained inspector model needs to operate on embedded hardware at the point of inspection instead of cloud infrastructure, this result has practical consequences. Lightweight deployment of solar defect classifiers is possible without catastrophic accuracy penalties, according to the comparatively modest performance differential between MobileNetV2 and heavier architectures like DenseNet121. This conclusion merits additional exploration in future work.

\subsection{Per-class evaluation}
The class-level precision, recall, and F1-score for ResNet101V2, the top-performing model on SolarFCD, are shown in Table \ref{tab:perclass_result}.
\begin{table}[h!]
\caption{Per-class classification results — ResNet101V2}
\label{tab:perclass_result}
\begin{tabular}{lllll}
\toprule
\textbf{Class}       & \textbf{Precision} & \textbf{Recall} & \textbf{F1-Score} & \textbf{Accuracy}               \\
\midrule
Healthy              & 0.892              & 0.889           & 0.887             & \multirow{4}{*}{\textbf{86.68}} \\
Surface\_Obstruction & 0.884              & 0.882           & 0.879             &                                 \\
Structural\_Fault    & 0.887              & 0.895           & 0.881             &                                 \\
Electrical\_Fault    & 0.883              & 0.889           & 0.890             &               \\
\bottomrule
\end{tabular}
\end{table}
All four categories show a remarkably consistent performance distribution, according to the per-class data, with precision, recall, and F1-score values falling within a small range of roughly 1.6 percentage points. Surface Obstruction records the lowest F1-score at 0.879, which is consistent with the visual ambiguity between light soiling and a clean panel surface under different illumination conditions. Electrical Fault achieves the highest F1-score at 0.890, slightly outperforming the Healthy class at 0.887. Given that missed structural faults carry the highest risk of module failure, Structural Fault achieves the maximum recall at 0.895, showing that the model is especially sensitive to structural damage signs. This is the most operationally desirable result.

\begin{figure}[h!]
    \centering
    \includegraphics[width=0.8\linewidth]{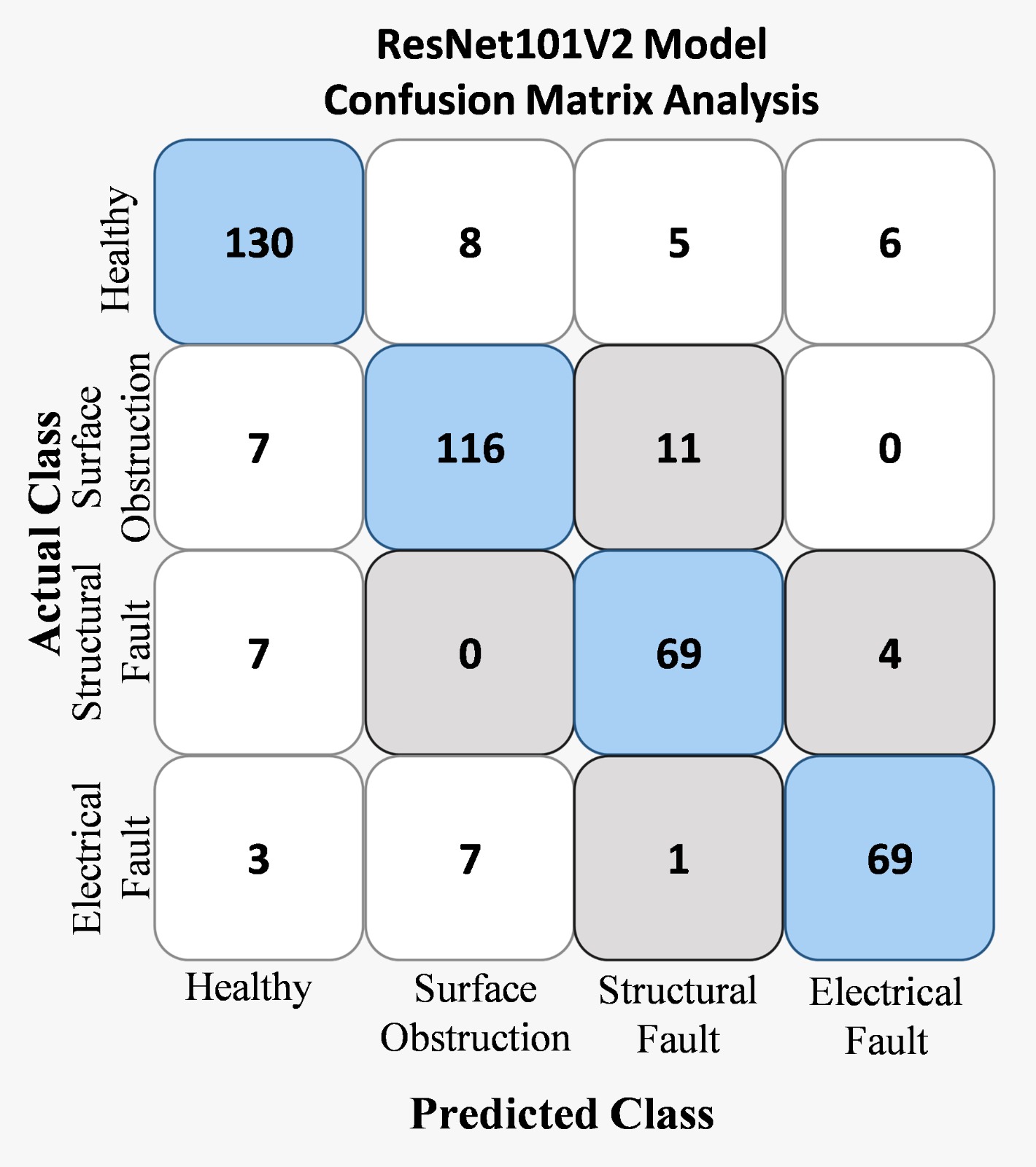}
    \caption{Confusion matrix of per-class performance}
    \label{fig:confusion_matrix}
\end{figure}

The confusion matrix of per-class performance for ResNet101V2 are shown in Figure \ref{fig:confusion_matrix}, which shows the distribution of accurate and incorrect predictions for each of the four classes. The confusion matrix shows that while cross-class confusion between structural fault and electrical fault is still low, indicating that the model has learned modality-appropriate discriminative features for the two most important defect categories, most misclassifications occur between Surface Obstruction and Healthy, reflecting the visual similarity between lightly soiled and clean panels.
\section{Usage Notes}
SolarFCD is appropriate for multi-class classification, transfer learning benchmarking, class imbalance research, and cross-modality generalization studies. It allows image-level classification across four defect categories: healthy, surface obstruction, structural fault, and electrical fault. Images should be resized to 224 x 224 pixels before training, ImageNet normalization should be used, and the given class weights should be utilized with cross-entropy loss to allow for residual imbalance; evaluation should report both overall accuracy and per-class metrics. Users should be aware that the training splits for structural fault and electrical fault contain augmented synthetic samples that can be identified by the augmentation flag in the metadata CSV; annotations are image-level only and do not support detection or segmentation tasks; and the thermal subset is derived from a single source, which limits environmental diversity. Any use of this dataset must cite this paper and the three source datasets.

\section*{Declarations}
\begin{itemize}
\item \textbf{Ethics approval and consent to participate.}
\item \textbf{Funding.} No funding
\item \textbf{Declaration of competing interest.} The authors declare that they have no known competing financial interests or personal relationships that could have appeared to influence the work reported in this paper.
\item \textbf{Consent for publication.} Not applicable
\item \textbf{Data availability.} Data is publicly available at: \href{https://cloud.dfki.de/owncloud/index.php/s/SZYo97gdwY2k92Z}{Dataset}
\item \textbf{CRediT authorship contribution statement.} Misbah Ijaz \& Saif Ur Rehman Khan: Conceptualization, Data curation, Methodology, Software, Validation, Writing original draft \& Formal analysis. Muhammed Nabeel Asim, Sebastian Vollmer \& Andreas Dengel: Conceptualization, Funding acquisition, Review. Abd Ur Rehman: Supervision, review \& editing. Arooj Zaib: alidation, Writing original draft.
\end{itemize}

\backmatter

\bibliography{sn-bibliography}

@article{johnson2021stabilisation,
  title={Stabilisation wedges: measuring progress towards transforming the global energy and land use systems},
  author={Johnson, Nathan and Gross, Robert and Staffell, Iain},
  journal={Environmental Research Letters},
  volume={16},
  number={6},
  pages={064011},
  year={2021},
  publisher={IOP Publishing},
  DOI={10.1088/1748-9326/abec06}
}

@article{zhang2024balancing,
  title={Balancing urban energy considering economic growth and environmental sustainability through integration of renewable energy},
  author={Zhang, Mei and Zhang, Danting and Xie, Tingfeng},
  journal={Sustainable Cities and Society},
  volume={101},
  pages={105178},
  year={2024},
  publisher={Elsevier},
  DOI={https://doi.org/10.1016/j.scs.2024.105178}
}

@techreport{iea_renewables_2025,
  author      = {{International Energy Agency (IEA)}},
  title       = {Renewables 2025},
  institution = {International Energy Agency},
  year        = {2025},
  address     = {Paris},
  url         = {https://www.iea.org/reports/renewables-2025}
}

@article{bamisile2025environmental,
  title={The environmental factors affecting solar photovoltaic output},
  author={Bamisile, Olusola and Acen, Caroline and Cai, Dongsheng and Huang, Qi and Staffell, Iain},
  journal={Renewable and sustainable energy reviews},
  volume={208},
  pages={115073},
  year={2025},
  publisher={Elsevier},
  DOI={https://doi.org/10.1016/j.rser.2024.115073}
}

@article{said2018effect,
  title={The effect of environmental factors and dust accumulation on photovoltaic modules and dust-accumulation mitigation strategies},
  author={Said, Syed AM and Hassan, Ghassan and Walwil, Husam M and Al-Aqeeli, N},
  journal={Renewable and Sustainable Energy Reviews},
  volume={82},
  pages={743--760},
  year={2018},
  publisher={Elsevier}, 
  DOI={https://doi.org/10.1016/j.rser.2017.09.042}
}

@article{malek2023design,
  title={Design and implementation of sustainable solar energy harvesting for low-cost remote sensors equipped with real-time monitoring systems},
  author={Malek, Kaveh and Rodr{\'\i}guez, Edgardo Ort{\'\i}z and Lee, Yi-Chen and Murillo, Joshua and Mohammadkhorasani, Ali and Vigil, Lauren and Zhang, Su and Moreu, Fernando},
  journal={Journal of Infrastructure Intelligence and Resilience},
  volume={2},
  number={3},
  pages={100051},
  year={2023},
  publisher={Elsevier},
  DOI={https://doi.org/10.1016/j.iintel.2023.100051}
}

@article{das2018metaheuristic,
  title={Metaheuristic optimization based fault diagnosis strategy for solar photovoltaic systems under non-uniform irradiance},
  author={Das, Saborni and Hazra, Abhik and Basu, Mousumi},
  journal={Renewable energy},
  volume={118},
  pages={452--467},
  year={2018},
  publisher={Elsevier},
  DOI={https://doi.org/10.1016/j.renene.2017.10.053}
}

@article{dhanraj2021effective,
  title={An effective evaluation on fault detection in solar panels},
  author={Dhanraj, Joshuva Arockia and Mostafaeipour, Ali and Velmurugan, Karthikeyan and Techato, Kuaanan and Chaurasiya, Prem Kumar and Solomon, Jenoris Muthiya and Gopalan, Anitha and Phoungthong, Khamphe},
  journal={Energies},
  volume={14},
  number={22},
  pages={7770},
  year={2021},
  publisher={MDPI},
  DOI={https://doi.org/10.3390/en14227770}
}

@article{wang2024locating,
  title={Locating the suitable large-scale solar farms in China's deserts with environmental considerations},
  author={Wang, Yimeng and Liu, Benli and Peng, Huaiwu and Jiang, Yingsha},
  journal={Science of The Total Environment},
  volume={955},
  pages={176911},
  year={2024},
  publisher={Elsevier},
  DOI={https://doi.org/10.1016/j.scitotenv.2024.176911}
}

@article{shaban2024detection,
  title={Detection and classification of photovoltaic module defects based on artificial intelligence},
  author={Shaban, Warda M},
  journal={Neural Computing and Applications},
  volume={36},
  number={27},
  pages={16769--16796},
  year={2024},
  publisher={Springer}
}

@article{hossain2025enhancing,
  title={Enhancing solar panel performance: A machine learning approach to dust detection and automated water sprinkle-based cleaning strategy},
  author={Hossain, Salman and Arika, All Mumtahina and Fahim, Iffat Nowshin and Uddin, Jamal and Ahmed, Ashik and Apon, Hasan Jamil and Hoque, Muhammad Arshadul},
  journal={Solar Energy},
  volume={287},
  pages={113240},
  year={2025},
  publisher={Elsevier},
  DOI={https://doi.org/10.1016/j.solener.2025.113240}
}

@article{jonathan2025multimodal,
  title={A multimodal deep learning approach for very short-term solar forecasts using sky images and historical numerical data},
  author={Jonathan, Anto Leoba and Bamisile, Olusola and Cai, Dongsheng and Ejiyi, Chukwuebuka Joseph and Nkou, Joseph Junior Nkou and Victor, Kombou and Ukwuoma, Chiagoziem C and Wei, Liu and Huang, Qi},
  journal={Renewable Energy},
  volume={255},
  pages={123774},
  year={2025},
  publisher={Elsevier},
  DOI={https://doi.org/10.1016/j.renene.2025.123774}
}

@inproceedings{mehta2024utilizing,
  title={Utilizing CNN-GAN for Enhanced Detection and Classification of Dust on Solar Panels},
  author={Mehta, Shiva and Kundra, Danish},
  booktitle={2024 5th International Conference on Smart Electronics and Communication (ICOSEC)},
  pages={915--919},
  year={2024},
  organization={IEEE},
  DOI={https://doi.org/10.1109/ICOSEC61587.2024.10722527}
}

@article{ejiyi2024maccom,
  title={MACCoM: A multiple attention and convolutional cross-mixer framework for detailed 2D biomedical image segmentation},
  author={Ejiyi, Chukwuebuka Joseph and Qin, Zhen and Ejiyi, Makuachukwu Bennedith and Ukwuoma, Chiagoziem and Ejiyi, Thomas Ugochukwu and Muoka, Gladys Wavinya and Gyarteng, Emmanuel SA and Bamisile, Olusola O},
  journal={Computers in Biology and Medicine},
  volume={179},
  pages={108847},
  year={2024},
  publisher={Elsevier}, 
  DOI={https://doi.org/10.1016/j.compbiomed.2024.108847}
}

@article{tella2025solar,
  title={Solar photovoltaic panel cells defects classification using deep learning ensemble methods},
  author={Tella, Hambal and Hussein, Alaa and Rehman, Shafiqur and Liu, Bou and Balghonaim, Adil and Mohandes, Mohamed},
  journal={Case Studies in Thermal Engineering},
  volume={66},
  pages={105749},
  year={2025},
  publisher={Elsevier},
  DOI={https://doi.org/10.1016/j.csite.2025.105749}
}

@article{araji2024utilizing,
  title={Utilizing deep learning towards real-time snow cover detection and energy loss estimation for solar modules},
  author={Araji, Mohamad T and Waqas, Ali and Ali, Rahmat},
  journal={Applied Energy},
  volume={375},
  pages={124201},
  year={2024},
  publisher={Elsevier},
  DOI={https://doi.org/10.1016/j.apenergy.2024.124201}
}

@article{guo2023accurate,
  title={Accurate and generalizable photovoltaic panel segmentation using deep learning for imbalanced datasets},
  author={Guo, Zhiling and Zhuang, Zhan and Tan, Hongjun and Liu, Zhengguang and Li, Peiran and Lin, Zhengyuan and Shang, Wen-Long and Zhang, Haoran and Yan, Jinyue},
  journal={Renewable Energy},
  volume={219},
  pages={119471},
  year={2023},
  publisher={Elsevier},
  DOI={https://doi.org/10.1016/j.renene.2023.119471}
}

@article{Pvmd,
  title={Photovoltaic module dataset for automated fault detection and analysis in large photovoltaic systems using photovoltaic module fault detection},
  author={Bello, Rotimi-Williams and Owolawi, Pius A and van Wyk, Etienne A and Du, Chunling},
  journal={Data in Brief},
  volume={57},
  pages={111184},
  year={2024},
  publisher={Elsevier},
  DOI={https://doi.org/10.1016/j.dib.2024.111184}
}

@misc{RGBdataset,
  author       = {P. Afroz},
  title        = {Solar Panel Images Clean and Faulty Images},
  year         = {2023},
  publisher    = {Kaggle},
  howpublished = {\url{https://www.kaggle.com/datasets/pythonafroz/solar-panel-images}},
  note         = {Accessed: 2026-04-25}
}

@misc{dust-dataset,
  author       = {H. S. Garladinne},
  title        = {Solar Panel Dust Detection},
  year         = {2023},
  publisher    = {Kaggle},
  howpublished = {\url{https://www.kaggle.com/datasets/hemanthsai7/solar-panel-dust-detection}},
  note         = {Accessed: 2026-04-25}
}

@article{chen2025open,
  title={Open data sets for assessing photovoltaic system reliability},
  author={Chen, Xin and Li, Baojie and Braid, Jennifer L and Byford, Brandon and Colvin, Dylan J and Glaws, Andrew and Jost, Norman and Pierce, Benjamin and Rabade, Salil and Springer, Martin and others},
  journal={Applied Energy},
  volume={395},
  pages={126132},
  year={2025},
  publisher={Elsevier}
}

\end{document}